\documentclass[10pt,twocolumn,letterpaper]{article}

\usepackage{wacv}
\usepackage{times}
\usepackage{epsfig}
\usepackage{graphicx}
\usepackage{amsmath}
\usepackage{amssymb}
\usepackage{subcaption}
\usepackage{booktabs}
\usepackage{float}
\usepackage[accsupp]{axessibility}
\usepackage{enumitem}

\makeatletter
\newcommand*\bigcdot{\mathpalette\bigcdot@{.5}}
\newcommand*\bigcdot@[2]{\mathbin{\vcenter{\hbox{\scalebox{#2}{$\m@th#1\bullet$}}}}}
\makeatother

\def\wacvPaperID{1306} 

\wacvfinalcopy 

\ifwacvfinal
\def\assignedStartPage{9876} 
\fi

\ifwacvfinal
\usepackage[breaklinks=true,bookmarks=false]{hyperref}
\else
\usepackage[pagebackref=true,breaklinks=true,colorlinks,bookmarks=false]{hyperref}
\fi

\ifwacvfinal
\else
\fi

\begin{document}

\title{Pose and Joint-Aware Action Recognition}

\author{
	Anshul Shah\textsuperscript{1} ~~~~
	Shlok Mishra\textsuperscript{2} ~~~~
	Ankan Bansal\textsuperscript{2} ~~~~
	Jun-Cheng Chen\textsuperscript{3} ~~~~ \\
	Rama Chellappa\textsuperscript{1} ~~~~
	Abhinav Shrivastava\textsuperscript{2} ~~~~\\
	\\
	\textsuperscript{1}Johns Hopkins University \ \ \ \ \ 
	\textsuperscript{2}University of Maryland, College Park \ \ \ \ \  \\
	\textsuperscript{3}Research Center for Information Technology Innovation, Academia Sinica \ \ \ \ \ \\ 
	{\small \tt \{ashah95,rchella4\}@jhu.edu~\{shlokm,ankan,abhinav2\}@umd.edu~pullpull@citi.sinica.edu.tw}
}
\maketitle

\ifwacvfinal
\thispagestyle{empty}
\fi


\begin{abstract}
Recent progress on action recognition has mainly focused on RGB and optical flow features. In this paper, we approach the problem of 
joint-based action recognition. Unlike other modalities, constellation of joints and their motion generate models with succinct human motion information for activity recognition. We present a new model for joint-based action recognition, which first extracts motion features from each joint separately through a shared motion encoder before performing collective reasoning. Our joint selector module re-weights the joint information to select the most discriminative joints for the task. We also propose a novel joint-contrastive loss that pulls together groups of joint features which convey the same action. We strengthen the joint-based representations by using a geometry-aware data augmentation technique which jitters pose heatmaps while retaining the dynamics of the action. We show large improvements over the current state-of-the-art joint-based approaches on JHMDB, HMDB, Charades, AVA action recognition datasets. A late fusion with RGB and Flow-based approaches yields additional improvements. Our model also outperforms the existing baseline on Mimetics, a dataset with out-of-context actions. 
\end{abstract}
\section{Introduction}

The task of action recognition has seen a lot of advances in recent times with improved spatio-temporal modeling, faster models and longer range temporal understanding. Most of the recent approaches in this area make use of raw RGB or dense optical flow as input features to reason about actions. In this paper, we approach the task of action recognition using joints and their motion. Unlike the commonly used dense optical flow, joints convey motion information succinctly, relying only on a sparse set of keypoints. Johansson's seminal work \cite{johansson1973visual} on `Moving Light Display' showed the importance of moving joints for human perception. In their experiments, bright spots were attached to joints of an actor dressed in black who was moving in front of a dark background. When the actor was not moving, the collection of spots was not helpful in discerning the action but when the person starts moving, the relative motion creates an impression of a person walking, dancing, etc. Further, many existing vision-based models tend to be biased by static context like scenes and objects \cite{Li2018RESOUNDTA, Li2019REPAIRRR,weinzaepfel2021mimetics}. Using joints helps to reduce these biases as it tends to be more robust to scene variations and explicitly captures only the human motion.

Recently, several interesting approaches for action recognition using human pose as a feature have been proposed. Approaches like PoTion\cite{choutas2018potion}, PA3D\cite{yan2019pa3d}, SIP-Net\cite{weinzaepfel2021mimetics} have shown impressive results. But, these approaches have certain limitations like overfitting on small datasets\cite{choutas2018potion}, requirement for access to multiple pose-modalities \cite{yan2019pa3d,weinzaepfel2021mimetics},  pose-tracking \cite{weinzaepfel2021mimetics}. Many of these approaches also require access to features from a pose extractor, making it difficult for the model to be applied to different off-the-shelf pose extractors. Most importantly, we note that all of these approaches do a collective joint-reasoning step right from the first layer. Reasoning about activities in videos using joints requires the model to integrate motion information from multiple joints. But we found that early fusion of joint information does not let the model learn good per-joint motion signatures and might make the model rely a lot on co-occurrence patterns which can lead to sub-optimal representations.

\begin{figure*}[ht!]
\centering
    \includegraphics[width=\linewidth]{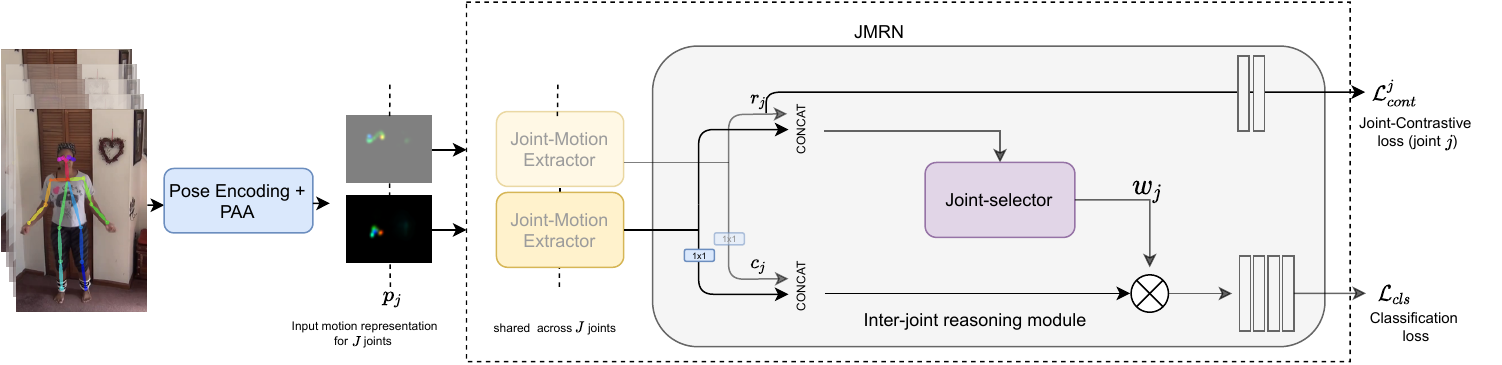}
\caption{The overall pipeline. We propose the joint-motion re-weighting network (JMRN) to better capture inter-dependencies between joints. This model captures motion information while learning to use the most discriminative joints. We first encode joint trajectories for each joint and augment them with the Pose-Aware augmentation to obtain the input motion representations $p_j,~~j \in [1,\cdots,J]$. The joint-motion extractor module is a shared encoder that learns information about each of the $J$ joints separately. This information is then forwarded to the inter-joint reasoning module, which learns a weight $w_j$ for each joint and reweighs the representation. In addition to the classification loss, we enforce a novel contrastive loss on the MLP-projected joint representations. The joint-contrastive loss is calculated for each joint separately. 
}
\vspace{-0.15in}
\label{fig:joint_assoc}
\end{figure*}

In this paper, we propose to alleviate the aforementioned issues through our approach Joint-Motion Reasoning Network (JMRN). First, in order to learn richer motion representations, we extract per joint motion information using a shared motion encoder. This is followed by a collective joint reasoning module to infer the final activity. 
Our approach of separating per-joint feature extraction and collective reasoning allows for additional constraints on the motion representations.
We propose a novel joint-contrastive loss which pulls together features from the same action for each joint. This loss enforces that the per-joint motion signatures are expressive of the action. Different from the usual contrastive learning setup \cite{chen2020simple,NEURIPS2020_d89a66c7} where the loss is applied to the last layer features, the proposed loss operates on the `mid-level' joint features. To improve the generalization capability of the model, we propose to use two geometry-aware data augmentation techniques. Fig.~\ref{fig:joint_assoc} shows the schematic of our model. Our proposed approach is not tied to any specific pose-extractor and does not require any pose tracking, which is difficult for in-the-wild videos. 

Using the proposed approach, we obtain an absolute accuracy improvement of 7.35\% on JHMDB \cite{Jhuang:ICCV:2013}, 3.9\% on HMDB\cite{Kuehne11}, 2.4 mAP on Charades \cite{Sigurdsson2016HollywoodIH}, and 1 mAP on AVA\cite{gu2018ava} over approaches which use pose heatmaps alone.
When combined with recent RGB and flow-based methods, our method yields additional improvements without retraining the backbone, demonstrating its effectiveness. We also evaluate our method on the Mimetics \cite{weinzaepfel2021mimetics} dataset. This dataset consists of out-of-context actions performed by mime artists. Evaluation on this dataset is meant to show the generalization capability of action recognition approaches. Standard RGB and flow-based models show dismal performance on this dataset. Our approach outperforms the baseline reported in \cite{weinzaepfel2021mimetics} by 1.5\% on top 1 accuracy without performing any tracking and only relying on joint heatmaps. Extensive analysis illustrates the importance of each proposed technique, and we often find that even our individual techniques outperform previous approaches. 

In summary, we make the following contributions:
\begin{itemize}
    \itemsep0em
    \item We propose a new model to improve reasoning over multiple human joints. This model learns motion representations for each joint independently, then uses an inter-joint reasoning module to combine all the motion cues from various joints to make a prediction. 
    \item Our novel joint-contrastive loss guides the joint representations to extract richer features. To the best of our knowledge, a contrastive learning loss on `mid-level' joint features has not been explored in the past.
    \item We propose a geometry-aware data augmentation method to deal with limited data and improve generalization.
    \item We obtain state-of-the-art results on JHMDB, HMDB, Charades and AVA when using the joints alone. The late fusion of our approach with RGB and flow-based models gives improvements showing the complementary nature of pose information. Our model also obtains an improvement on Mimetics\cite{weinzaepfel2021mimetics}, a dataset with out-of-context actions using only pose heatmaps and without any tracking. 
\end{itemize}
\section{Related Works}
\textbf{Action Recognition from Videos.}
\cite{simonyan2014two} was amongst the first to successfully apply deep networks for action recognition that relied on RGB and optical flow. \cite{carreira2017quo} improved performance on action recognition task by inflating weights from 2D CNNs.  Some recent advances in learning improved spatio-temporal representations \cite{wang2016temporal,sigurdsson2017asynchronous,wang2018non,yue2018compact,cherian2017generalized,wu2019long,Kalfaoglu2020LateTM,feichtenhofer2019slowfast} have led to better performance in standard action recognition benchmarks. 
But, studies \cite{Li2018RESOUNDTA,Li2019REPAIRRR,weinzaepfel2021mimetics} show that RGB and flow-based models capture a lot of dataset biases. Unlike RGB and flow, human joints and their motion offer a succinct representation of the underlying activities and is less susceptible to dataset biases. To this end, we present an approach for action recognition using pose information.

\textbf{Pose-based Action Classification.}
\cite{Jhuang:ICCV:2013} showed that pose features perform much better than low/mid features and thus can act as discriminative cues for action recognition. Strides of improvements in pose estimation from images and videos \cite{cao2018openpose,Luo2018LSTMPM, Pfister2015FlowingCF,Toshev2014DeepPoseHP,fang2017rmpe,tompson2014joint,ke2018multi, carreira2016human, wei2016convolutional, yang2017learning, sun2019deep,moon2019posefix} have enabled extraction of pose information from in-the-wild videos for the task of action recognition. 

In P-CNN~\cite{CheronLS15}, the authors extract appearance and flow features conditioned on the human pose. Similarly, chained multi-stream networks \cite{zolfaghari2017chained} combines the most important visual cues from pose, motion and RGB. A Multi-task deep learning framework was proposed in \cite{luvizon20182d} for using visual information and joints to classify activities. A recurrent pose-attention network was introduced in \cite{DuW017} to learn the spatio-temporal features by attending to joints and extracting visual features. \cite{cao2016action} also extracts visual features near joints to obtain video descriptors. However, handling multiple people becomes difficult with such a method. These methods require access to ground-truth keypoints during training, making it difficult to train on videos without these annotations. A rank pooling-based approach was proposed in \cite{liu2018recognizing} to pool information from the evolution of pose-heatmaps. 

Most related to our method, \cite{choutas2018potion} proposed to represent the pose evolution as a compact representation and used it in a deep framework to obtain improvements for action recognition in real-world videos. \cite{yan2019pa3d} improved upon this by using Part Affinity Fields \cite{Cao2017RealtimeM2} and visual features from the pose extraction pipeline along with a temporal pooling module. In contrast, we do not use these features and work on joint heatmaps alone, which is a common representation used in all pose extractors and shows improvements without using extra modalities. STAR-Net \cite{McNallyWM19}, re-projects pose-estimation features in space and time to build an end-to-end system for action recognition. While they use clips, we work with the entire video. Further, in addition to demonstrating better results, our approach is modular and can work with any pose estimation model. DynaMotion \cite{asghari2020dynamic} proposed using a dynamics-based encoder which encodes input sequences using a structured dictionary. Similar to \cite{choutas2018potion}, they reason on all joints together. By extracting motion information from each joint independently before performing inter-joint reasoning helps our model extract richer motion representation which leads to superior results. This way of modeling also naturally allows us to use additional supervision on joint motion representations which leads to additional gains. 

Another line of work exploits the skeletal structure of humans to classify activities \cite{vemulapalli2014human,wei3d,song2017end,yan2018spatial,shi2019two}. Accurate 3D poses can be obtained using depth cameras, but it is not viable in general settings and obtaining 3D pose from unconstrained videos is still a very difficult problem. Hence we do not focus on these approaches in our work.
\section{Method}
In this section we first describe how we obtain the input per-joint motion trajectories for our approach, followed by our proposed approach

\textbf{Pose extraction and encoding pose evolution} In this paper, we focus on action recognition from videos using the evolution of human pose as a cue. Given a video of $T$ frames, we first run a pose detector on each frame of the video to obtain heatmaps $h_j^t \in \mathbb{R}^{H \times W},~t = 1, \dots, T$ for the $j^{\text{th}}$ joint, $j = 1, \dots, J$. The heatmaps intuitively denote the location estimate of each joint for every frame.  
Following \cite{choutas2018potion}, we temporally aggregate the heatmaps using a weighted sum. Given the temporally stacked joint heatmaps $C \in \mathbb{R}^{T\times J\times H \times W}$, we aggregate the heatmaps to give a representation $P' = \sum_t{C[t] o[t]}$ using a (fixed) piece-wise linear weighing function $o$ where $P' \in \mathbb{R}^{J\times H\times W}$. We use three different weighting functions described in~\cite{choutas2018potion}, giving the final motion representation of shape $p_j \in \mathbb{R}^{3\times H\times W}$.
The advantage of using this approach is that it lets us encode the whole video without sampling, and also helps in reasoning about the global structure more effectively using CNNs. This module returns joint-motion trajectories which are used by our model for action recognition. 

\subsection{The Proposed Model and Training}

Next, we introduce the structure of our JMRN which captures the joint correlations. Previous approaches used all joints concatenated together as input to the neural network. This can lead to the model not exploiting useful motion information from individual joints and relying on specific spatial arrangements of these found in the dataset. We argue that we can learn improved representations by independently extracting motion information from each joint using a shared module before combining the cues for inter-joint reasoning.  
Our next observation is that, for an activity, some joints might have more discriminative information than others. Some joints might act as `distractors' to the training process and only provide redundant or noisy information. We can learn enhanced features by using information from more discriminative joints.
Our model (Fig.~\ref{fig:joint_assoc}), JMRN is designed to solve these issues and learns effective representations for joint-based action recognition. The model consists of two modules: 
The first module extracts information from all joints separately. The second module is an inter-joint reasoning module that is conditional on the input and learns to weigh the features from the various joints before reasoning over altogether. 
The joint-selector module is similar in spirit to Squeeze-and-Excitation block \cite{hu2018squeeze}, but with a key difference that instead of operating on higher-level representations of the entire input, our representations contain motion information about that joint alone. This makes our selector module naturally interpretable. Finally, we stack the weighted motion representations and feed them to the inter-joint reasoning module. This module generates final logits by performing collective reasoning over all joints.

\noindent\textbf{Joint-Motion Extraction.} 
The joint-motion extraction module is a Siamese network that extracts information from each joint separately. 
The input to the network is the motion representation for the $j^{th}$ joint, $p_j \in \mathbb{R}^{3\times H\times W}$ 
The module generates a representation $r_j \in \mathbb{R}^{256\times H\times W},~j=1, \dots ,J$ which is then used by the joint-selector module. In addition, we also generate a compressed representation $c_j \in \mathbb{R}^{c_{dim}\times H\times W}$ which is obtained by passing $r_j$ through a $1 \times 1$ convolution layer to reduce the number of channels.  

\noindent\textbf{Inter-Joint Reasoning.} The joint-selector module pools information from all joint-motion representations $r_j$ to obtain weights that are used to modulate the representations $c_j$. Specifically, we concatenate all the joint representations and apply a $1 \times 1$ convolution. This is followed by an average pooling operation which generates a feature of dimension 256. A linear layer followed by sigmoid activation is used to generate the weights $w_j$ . 
Finally, we weigh the compressed representations $c_j$ by the obtained weights of $w_j$ and concatenate them. We use a $1 \times 1$ convolution to reduce the large dimension of the input ($J \times c_{dim}$) and follow this with two convolutional layers and an FC layer to give the final class logits. \\
\noindent\textbf{Joint-Contrastive Loss}
The classification loss will naturally try to cluster final feature representations from the same class together. Extracting per-joint features before collective joint reasoning allows enforcing additional constraints on the joint features. Consider an instance of a person `running'. While the standard classification loss enforces that concatenated features from instances of running lie close to each other, we can learn improved motion features by enforcing consistencies at the joint level. Specifically, our joint-contrastive loss ensures that per-joint features for instances of `running' lie closer to each other than per-joint features of `push-up'. We enforce this constraint by employing contrastive learning loss~\cite{oord2018representation,chen2020simple,NEURIPS2020_d89a66c7}. Different from prior works, our loss operates on `mid-level' joint motion features. Further, we apply this loss for each joint separately. Our positives come from augmented examples of the same instance and other instances of the same activity while the negatives are other remaining instances from the batch. We first project the joint features $r_j$ through an MLP and normalize them to obtain features $\boldsymbol{z_j}$ which lie on a unit hypersphere. Let the corresponding label for the instance be denoted by $\boldsymbol{y}$. The joint-contrastive loss is then defined by:
\begin{align}
\mathcal{L}_{cont} &= \sum_{j=1}^{j=J} \mathcal{L}_{cont}^{j} 
\end{align}

\begin{align}
      \mathcal{L}_{cont}^{j} = \sum_{i\in B}\frac{-1}{|P(i)|}&\sum_{p\in P(i)}\log{\frac{\text{exp}\left(\boldsymbol{z}_j^i\bigcdot\boldsymbol{z}_j^p/\tau\right)}{\sum\limits_{a\in B \setminus {i} }\text{exp}\left(\boldsymbol{z}_j^i\bigcdot\boldsymbol{z}_j^a/\tau\right)}}
\end{align}

\noindent where $B$ are the instances in the batch, $z_j^i$ denotes the projected features for the joint $j$ of instance $i$ and $P(i)\equiv\{p\in B \setminus {i}:\boldsymbol{y}^p=\boldsymbol{y}^i\}$, is the set of all instances which share the same label as instance $i$ (positives).  $\tau$ is the temperature parameter. To the best of our knowledge, a contrastive learning loss on joint features to improve representations has not been explored in the past.
\subsection{Pose-Aware Data Augmentation (PAA)}
Data augmentation, like flipping the representation, has been shown to be useful in previous approaches~\cite{choutas2018potion,yan2019pa3d}. 
We propose a data augmentation strategy which can use the geometric structure of the human joints. Our pose-aware data augmentation is a cascade of two operations. The first operation is a random global jitter of the entire input representation. This helps the network to learn that the action is not strongly dependent on the global position of joint trajectories and ensures that the relative position among the different joints is not modified. 
Another benefit is that it regularizes training by sometimes pushing joints out of the frame which emulates joints not being visible in the scene. 
In the second operation, we divide the pose into six groups for joints corresponding to the Head:~\{Nose,~REye,~LEye,~LEar,~REar\},
Torso:~\{RHip,~LHip,~Neck\},  
Left Hand:~\{LShoulder,~LElbow,~LWrist\},  Right Hand:~\{RShoulder,~RElbow,~RWrist\},  Left Leg:~\{LKnee,~LAnkle\}, and  Right Leg:~\{RKnee,~RAnkle\}.
Each joint inside a group is randomly jittered by the same amount. As our first step involves a global motion, we do not jitter the joints in the torso group. This strategy amounts to adding random spatial noise but is pose-aware in the sense that all joints do not move by a different random amount. For example, the motion representations corresponding to eyes, ears and nose are jittered by the same amount. We represent our augmentation by parameters $\beta,\gamma$, which represents the maximum amount of random jitter for each of the two strategies, respectively. The same amount of jitter is used for all frames of a video to preserve the geometric integrity of the action.
Similar to pose-aware translation, we also experiment with pose-aware rotation which performs comparably. We include those experiments in the supplementary material. We note that while some of these data augmentation techniques have been explored for pose-estimation, these have not been investigated for the task of action recognition which is a very different task conceptually and might require the model to learn different kind of invariances. 

\section{Experiments}
In this section, we present experimental results to show the effectiveness of the proposed approach. 
Following prior works, we use HMDB \cite{Kuehne11}, JHMDB \cite{Jhuang:ICCV:2013}, Charades \cite{Sigurdsson2016HollywoodIH}, AVA\cite{gu2018ava} and Mimetics \cite{weinzaepfel2021mimetics} datasets for our experiments. 
\newline
\textbf{HMDB~\cite{Kuehne11}} consists of 6,766 video clips from fifty-one action classes. Each video clip is trimmed and corresponds to a single action. There are 3,570 training videos and 1,530 videos for validation in each of the three splits.
\\
\textbf{JHMDB~\cite{Jhuang:ICCV:2013}} is a subset of HMDB that contains 928 short videos from twenty-one action classes. The dataset comprises of $\sim$660 training videos and $\sim$268 validation videos for each of the three splits. 
\newline
\textbf{Charades~\cite{Sigurdsson2016HollywoodIH}} is a more recent dataset that contains daily-life videos instead of videos from movies or Youtube. The dataset has 157 action classes with a total of 9,848 untrimmed videos, of which 7,981 videos are for training and 1,863 are for validation. 
\\
\textbf{AVA Actions~\cite{gu2018ava}} is a large-scale dataset which evaluates video recognition models on spatio-temporal action localization. One frame every second is annotated with actors and each actor might be involved in multiple actions. We use the AVA v2.1 set which consists of 211k training segments and 57k validation segments. Following previous works, we evaluate mAP at frame-level IoU threshold of 0.5 on sixty actions which have at least twenty-five instances in the validation set.
\\
\textbf{Mimetics~\cite{weinzaepfel2021mimetics}} is a test set for fifty actions from the Kinetics dataset. It consists of 713 videos collected from Youtube that have mimed actions. Due to the small size of this dataset, it is only used for testing a model that is trained with Kinetics or its subset. Unlike most common action recognition datasets, this dataset consists of minimal scene and object biases and hence can be used to evaluate a method for out-of-context actions.

\subsection{Implementation Details}
\label{implementation_details}
For all the datasets we extract pose using an off-the-shelf pose detector \cite{cao2018openpose} which gives heatmaps per frame for the video. We use the COCO trained pose extractor which returns 18 joints + background. 
We train the models using the Adam optimizer with a learning rate drop on accuracy plateau. We used cross-entropy for single label classification tasks and binary cross-entropy for multi-label tasks. Models are trained with the loss function $\mathcal{L}_{cls} + \lambda \mathcal{L}_{cont}$ where $\lambda$ is chosen empirically. For multi-label tasks, two instances are considered positives if they share any of the labels. We used our own implementation of PoTion (baseline) since the official implementation is not publicly available. 

\subsection{Results}
We show the results of our proposed approaches in Table  \ref{tab:improvements_table}. The baseline model is PoTion \cite{choutas2018potion}. The baseline model stacks heatmaps from all joints and uses the combined representation to recognize actions. The proposed model, JMRN first extracts motion information from each joint using a shared module before reweighing this information for inter-joint reasoning. This gives consistent gains on all datasets. Adding the novel joint-contrastive loss to enforce feature consistency at the joint-level brings in additional improvements. Our proposed pose-aware data augmentation gives a considerable improvement to the baseline model and our JMRN model over all datasets.

\begin{table*}
\centering
\small\renewcommand{\tabcolsep}{6pt}
\renewcommand{\arraystretch}{1.2}
\caption{Improvements using the proposed approach. We compare our model agains the baseline (PoTion\cite{choutas2018potion}). Our proposed model outperforms the baseline by a large margin. Use of the proposed data augmentation step (PAA) improves the JMRN model consistently across the three datasets. Finally, using the novel joint-contrastive loss in addition to the classification loss gives further improvements.}
  \vspace{-0.05in}
\begin{tabular}{@{}llccc@{}}  
\toprule
 & \textbf{Approach} & \textbf{JHMDB-1} & \textbf{HMDB-1} & \textbf{Charades}\\
\toprule
& Baseline (PoTion) & $59.44$  & $42.04$ & $13.54$ \\
\midrule
&JMRN &    $66.70$ &  $48.71$   &  $15.00$\\
\textbf{Proposed Model} & + PAA  & $69.81$   & $52.02$ &  $15.79$\\
& + PAA + Joint-Contrastive Loss  & \textbf{71.08}  &  \textbf{54.05} & \textbf{16.2}\\
\bottomrule
\end{tabular}
\label{tab:improvements_table}
\vspace{-0.1in}
\end{table*}

\begin{table}[t]
\centering
\small
\renewcommand{\arraystretch}{1.2}
\renewcommand{\tabcolsep}{8pt}
\caption{Comparison with state-of-the-art on J-HMDB and HMDB datasets. We report mean-per class accuracy averaged over 3 splits. We obtain an improvement of +7.35\% on JHMDB and +3.9\% on HMDB when using pose heatmaps alone. Note that the best results for \cite{yan2019pa3d} make use of extra modalities and additional networks to process these. }
  \vspace{-0.05in}
\resizebox{\linewidth}{!}{ 
\begin{tabular}{llcccc}  
\toprule
\textbf{Method}  & \textbf{Features Used} & \textbf{HMDB} & \textbf{JHMDB}\\
\midrule
PA3D~\cite{yan2019pa3d} & P + CF + PAF & 55.3* & 69.5*\\
\midrule
\midrule
Potion~\cite{choutas2018potion} & P & 43.7 & 57.0\\
PA3D~\cite{yan2019pa3d} & P & 47.8 & 60.1 \\
PA3D~\cite{yan2019pa3d} & P & 50.3* & 61.2* \\
STAR-Net~\cite{McNallyWM19} & CF & - & 64.3 \\
EHPI~\cite{ludl2019simple} & P & - & 60.5 \\
DynaMotion~\cite{asghari2020dynamic} & P & 49.1$^{+}$ & 60.2$^{+}$ \\
SIP-Net~\cite{weinzaepfel2021mimetics} & CF & 51.2 & 62.4 \\
\midrule
Ours & P & \textbf{54.2} & \textbf{68.55}\\
\bottomrule
\label{tab: jhmdb_hmdb_pose_best}
\end{tabular}
}
Models with * use difference features along with standard heatmaps and use an ensemble pose model. + are results on first split alone.
P - pose heatmaps, PAF - Part affinity Fields \cite{Cao2017RealtimeM2}, CF - Features from the pose extractor CNN.
\end{table}

\begin{table}[t]\centering
\small
\renewcommand{\arraystretch}{1.2}
\renewcommand{\tabcolsep}{8pt}
\caption{Classification mAP on the evaluation set of Charades dataset. We obtain an absolute improvement of 1.55 mAP over PA3D which uses PAF \cite{Cao2017RealtimeM2} and convolutional features apart from heatmaps. We also show results on combination with RGB (R) and Flow-based (F) approaches which gives further gains.}
  \vspace{-0.05in}
\resizebox{\linewidth}{!}{  
\begin{tabular}{@{}llc@{}}  
\toprule
\textbf{Method}  & \textbf{Features} & \textbf{mAP}\\
\midrule
2Stream \cite{simonyan2014two}       & R + F & 11.9\\   
Asyn-TF \cite{sigurdsson2017asynchronous} & R + F & 22.4 \\ 
I3D \cite{carreira2017quo}  & R & 32.72$^{\dagger}$ \\
GCN \cite{wang2018videos}  & R & 36.2\\
NL I3D \cite{wang2018non}   & R & 37.5 \\
R50-I3D-NL \cite{wu2019long} & R & 38.29$^\#$ \\
R101-I3D-NL LFB \cite{wu2019long} & R & 42.5$^\#$ \\
PoTion \cite{choutas2018potion} & P & 13.54$^\dagger$\\
Potion + R101-NL-LFB  & P + R & 42.84  \\
PA3D \cite{yan2019pa3d} & P + CF + PAF & 13.8 \\
PA3D \cite{yan2019pa3d} + GCN + I3D + NL-I3D & P + CF + PAF + R  & 41.0 \\
\midrule
Ours & P & \textbf{16.2} \\
Ours + R101-NL-LFB & P + R & \textbf{43.23} \\
\bottomrule
\label{tab: experiments with sota_charades}
\end{tabular}
}

  \vspace{-0.1in}
$\dagger$ denotes results we have reproduced. \# using official implementation. P - pose heatmaps, PAF - Part affinity Fields \cite{Cao2017RealtimeM2}, CF - Features from the pose extractor CNN. 
  \vspace{-0.1in}
\end{table}

\begin{table}[t]
\centering
\small
\renewcommand{\arraystretch}{1.2}
\renewcommand{\tabcolsep}{8pt}
\caption{Comparison with fusion over recent state-of-the-art on HMDB and JHMDB dataset. We report mean-per class accuracy averaged over 3 splits. The improvement through model fusion shows that our approach is complementary to RGB (R) and flow-based (F) models.}
  \vspace{-0.05in}
  
\resizebox{\linewidth}{!}{
\begin{tabular}{lcc}
\toprule
\textbf{Method}  & \textbf{Modality} & \textbf{HMDB}\\
\toprule
2Stream~\cite{simonyan2014two} & R + F & 59.4 \\
TSN~\cite{wang2016temporal} & R + F & 69.4  \\
S3D~\cite{xie2018rethinking} & R + F & 75.9 \\
R(2+1)D~\cite{tran2018closer} & R + F & 78.7  \\
I3D~\cite{carreira2017quo}  & R + F &  81.09$^\dagger$  \\
EvaNet~\cite{piergiovanni2019evolving} & R + F & 82.3 \\
PA3D~\cite{yan2019pa3d} + I3D~\cite{carreira2017quo}  & P + PAF + CF + R + F &  82.1$^*$\\
ResNext101 BERT ~\cite{Kalfaoglu2020LateTM}  & R + F &  83.76$^\dagger$\\
\midrule
Ours + I3D & P + R + F  & 82.33 \\
Ours + ResNext101 BERT & P + R + F  & \textbf{84.53}  \\
\bottomrule
\label{tab: experiments with sota hmdb2}
\end{tabular}
}
  \vspace{-0.05in}
\resizebox{\linewidth}{!}{
\begin{tabular}{lcc}  
\toprule
\textbf{Method}  & \textbf{Modality} & \textbf{JHMDB}\\
\toprule
I3D~\cite{carreira2017quo}  & R + F & 86.8$^\dagger$\\
P-CNN \cite{CheronLS15}       & R + F & 61.1\\ 
P-CNN + IDT \cite{CheronLS15}       & R + F & 71.4 \\    
Action Tubes \cite{Gkioxari_2015}  & R + F &  62.5\\   
MR-TS R-CNN \cite{Peng2016MultiregionTR}& R + F  &  71.1 \\
GRP + IDT \cite{cherian2017generalized} & R + F & 74.3 \\
KRP + IDT \cite{cherian2018non} & R + F & 74.2 \\
Chained MultiStream~\cite{zolfaghari2017chained}  & P + R + F & 76.1\\
PoTion\cite{yan2019pa3d} + I3D ~\cite{carreira2017quo}  & P + R + F & 85.5\\
PA3D\cite{yan2019pa3d} + RPAN ~\cite{Du2017RPANAE}  & P + PAF + CF + R + F & 86.1\\
\midrule
Ours + I3D & P + R + F  &  \textbf{88.36} \\
\bottomrule
\label{tab: jhmdb_hmdb_fusion_best}
\end{tabular}
}
$\dagger$ denotes results we have reproduced. Models with * use difference features along with standard heatmaps and use separate models for all. P - pose heatmaps, PAF - Part affinity Fields \cite{Cao2017RealtimeM2}, CF - Features from the pose extractor CNN.
\end{table}

\noindent\textbf{Comparison with prior Pose-based Methods.}
In Tables \ref{tab: jhmdb_hmdb_pose_best} and \ref{tab: experiments with sota_charades}, we compare our approach to other state-of-the-art approaches for the various datasets. The proposed approach performs better than other approaches that use only joint heatmaps. Our approach is more general, as unlike PAFs and CNN features, heatmaps can be obtained from any pose extractor. Further, unlike our method, PA3D's model uses additional inputs (PAF, CNN features, and temporal difference of features) and make use of ensembles of models. Our superior results without any of these bells and whistles show the benefit of our approach.

\textbf{Fusion with State-of-the-Art Approaches.}
While joint motion is a strong cue for human activities, many of the current datasets have activities that are strongly dependent on objects and scenes. Our pose stream can still offer complementary motion information for such datasets and lead to an improved reasoning of activities. 
Na\"ively averaging logits is sub-optimal for fusion since the logits are generated by models which are trained separately and might have a different effective range. We use a very simple learnt fusion scheme to combine the different modalities. Specifically, we learn a single scalar weighing parameter for each of the $M$ modalities to be fused and $M$ for each class for multi-label problems. 
This approach does not require backpropagation through the backbone and hence can be used as a quick post-training step with extracted logits. 
Fusing the scores with the best model, we obtain improvements on all datasets as shown in Tables \ref{tab: jhmdb_hmdb_fusion_best} and \ref{tab: experiments with sota_charades}. These results show that our model extracts joint information which is complementary to RGB and Flow. 
For HMDB and JHMDB, we fuse with I3D \cite{carreira2017quo}, and additionally ResNeXt101-BERT \cite{Kalfaoglu2020LateTM} which is the current state-of-the-art for HMDB. For Charades, we perform fusion with Long-Term Feature banks (LFB) \cite{wu2019long}. We see that our model enables complementary gains throughout. 
We found that on HMDB-1 test set, fusion of JMRN over ResNeXT101-BERT\cite{Kalfaoglu2020LateTM} was particularly useful for \texttt{jump}~(13\%), \texttt{turn}~(+7\%), \texttt{draw-sword}~(+7\%) and \texttt{run}~(+7\%).  
On Charades, we reap the largest absolute improvements on \texttt{standing up}~(+11\%), \texttt{sitting down}~(+10\%) and \texttt{washing a mirror}~(+10\%) over the LFB model. These actions are strongly tied to human motion and the improvements justify the use of a pose stream in addition to RGB and optical flow. We believe that jointly training these models would lead to further improvements wherein each stream can specialize for different classes, but we leave that to future work.

\noindent\textbf{Results on AVA Actions~\cite{gu2018ava}}
To show the advantage of using a pose-stream on large-scale datasets, we perform experiments on the AVA Actions dataset. We compare our approach with the best publicly available SlowFast \cite{feichtenhofer2019slowfast} model for this split. For fair comparison, we use bounding boxes provided by SlowFast \cite{feichtenhofer2019slowfast}. Pose is extracted for each frame in the clip and bounding box information is used to crop per-person pose information across the clip. These are then used to obtain the pose encoding. Since some actions involve multi-person context information, we also append the pose encoding of other people in the clip as separate channels. Our results are shown in Table~\ref{tab: ava_results}. We obtain an improvement over the strong RGB-based baseline using late fusion with JMRN's pose stream. 
It is particularly noteworthy that, unlike the SlowFast model which uses Kinetics-600 to pretrain the model, we train our model from scratch and our pose-only model has a comparable performance to the I3D RGB model. 

\begin{table}
\centering
\caption{Results on the validation set of the AVA v2.1 dataset. Results on this dataset demonstrate that the approach works on a large scale spatio-temporal action localization dataset. We see a significant improvement over the reproduced baseline. Further, we see that JMRN trained from scratch gives performance close to I3D-RGB model which was pretrained using large-scale Kinetics-400 dataset justifying the use of joint trajectories for action recognition. Fusion with a RGB-based model gives further gains.}
\resizebox{\linewidth}{!}{
\begin{tabular}{lccc}  
\toprule
Method  & Pretraining & Modality & mAP \\
\midrule
I3D \cite{carreira2017quo} & Kinetics-400 & RGB & 14.5 \\
I3D \cite{carreira2017quo} & Kinetics-400 & RGB + Flow & 15.6 \\
ACRN \cite{sun2018actor} & Kinetics-400 & RGB + Flow & 17.4 \\
I3D \cite{carreira2017quo} & Kinetics-600 & RGB & 21.9 \\
SlowFast-R101-NL \cite{feichtenhofer2019slowfast} & Kinetics-600 & RGB & 28.06$^\dagger$ \\
PoTion\cite{choutas2018potion} & None & Pose & 13.1 \\
\midrule
Ours & None & Pose & 14.1 \\

\midrule
Ours + SlowFast-R101-NL &  & Pose + RGB &  \textbf{28.4}\\
\bottomrule
\label{tab: ava_results}
\end{tabular}
}
$\dagger$ denotes results we have reproduced using the publicly available model. 
\end{table}

\noindent\textbf{Results on Mimetics dataset}
Mimetics \cite{weinzaepfel2021mimetics} introduced a test set for a subset of fifty actions found in the Kinetics dataset. The proposed dataset is collected from Youtube but consists of mimed actions and can be used to evaluate the performance of models on out-of-context actions and thus evaluate their generalization capability. Pose can be a strong cue in such cases as the videos have minimal scene and object biases evident from the observation that RGB/Flow-based models have poor performance on this dataset \cite{weinzaepfel2021mimetics}. Table \ref{tab: mimetics} lists our results. We obtain improvements on all previously reported metrics over SIP-Net \cite{weinzaepfel2021mimetics} while using only pose heatmaps and without relying on any tracking - further justifying our approach. Our approach also outperforms the pose-only model of \cite{Moon2020IntegralActionPF} which uses a more accurate pose extractor and we obtain 2.7\% improvement on top-5. 

\begin{table}[h!]
\centering
\small
\renewcommand{\arraystretch}{1.2}
\renewcommand{\tabcolsep}{6pt}
\caption{Mimetics experiments. Results when models are trained on the 50 Kinetics classes that intersect with Mimetics. We obtain an improvement over SIP-Net\cite{weinzaepfel2021mimetics} while not requiring any tracking and using pose heatmaps alone.}
\resizebox{0.6\linewidth}{!}{
\begin{tabular}{@{}lcccc@{}}  
\toprule
Method & top 1 & top 5 & mAP\\
\midrule
SIP-Net \cite{weinzaepfel2021mimetics} & 25.1 & 51.4 & 38.3\\
Ours & \textbf{26.6} & \textbf{52.7} & \textbf{40.0}\\
\bottomrule
\label{tab: mimetics}
\end{tabular}}
\centering
\small
\renewcommand{\arraystretch}{1.2}
\renewcommand{\tabcolsep}{6pt}
\caption{JMRN \vs Deeper baseline. Our model outperforms a deeper version of the model proposed in \cite{choutas2018potion} thus showing the effectiveness of our approach. }
\resizebox{0.8\linewidth}{!}{ 
\begin{tabular}{@{}lccc@{}}  
\toprule
 & JHMDB-1 & HMDB-1 & Charades\\
\toprule
Deeper Baseline & $66.38$ & $49.25$ &  $14.88$ \\
Ours & $\mathbf{71.08}$ & $\mathbf{54.05}$ & $\mathbf{16.2}$ \\
\bottomrule
\label{tab: more_params}
\end{tabular}
}
\vspace{-0.2in}
\end{table}

\subsection{Additional Analyses}

\noindent\textbf{JMRN \vs Deeper Baseline.} 
Our data augmentation strategy and model design helps us train deeper models and learn improved representations without over-fitting. To verify that our model leads to superior results compared to baseline due to design choices and not increased number of parameters, in Table \ref{tab: more_params} we experiment with a deeper version of the baseline model (PoTion). We add three convolutional layers to the baseline model, increasing the number of parameters in the baseline to match that of our model. For a fair comparison, we make use of our data augmentation scheme in the deeper baseline model too. The deeper baseline still performs worse than our proposed model, thus validating our proposed approach. 

\begin{table}
\centering
\small
\renewcommand{\arraystretch}{1.2}
\renewcommand{\tabcolsep}{6pt}
\caption{Effect of sharing parameters of the motion extractor module. The non-Siamese model performs worse than our proposed model.}
\vspace{-0.05in}
\resizebox{0.7\linewidth}{!}{
\begin{tabular}{@{}lccc@{}}  
\toprule
 & JHMDB-1 & HMDB-1 & Charades\\
\toprule
Non-Siamese & $65.98$ & $51.66$ & $14.93$\\
Ours & $\mathbf{71.08}$ & $\mathbf{54.05}$ & $\mathbf{16.2}$ \\
\bottomrule
\label{tab:non_siamese}
\end{tabular}}
\vspace{-0.2in}
\end{table}
\noindent\textbf{Comparison with a Non-Siamese Network.} Parameters of our motion extractor network are shared across all joints. Here we show the results when we do not share these parameters and instead learn separate motion extractors for each joint. To maintain a similar number of parameters as the baseline, the motion extractors are made shallower than in JMRN. Table \ref{tab:non_siamese} shows the results of these experiments. This model performs worse than our proposed JMRN model and shows the importance of sharing the parameters.

\begin{figure}
  \includegraphics[width=0.3\textwidth]{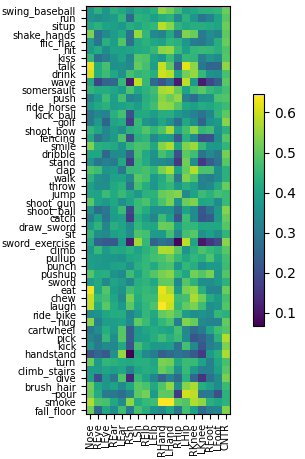}
  \vspace{-0.1in}
  \centering
\caption{Visualization of input conditional weights
$w_i$ for HMDB-1 test set. We show the average weights
for each class. Our approach of having a joint-selector as a part of the model allows for easy interpretation of joint importance for various actions. For example, we see that actions like `eat' and `drink' have a high dependence on hands.}
  \label{fig:viz_heatmap}
  \vspace{-0.2in}
\end{figure}

\noindent\textbf{What is the model learning?}
Since we first extract joint-motion signatures separately before fusing, we can attribute the weights in the selector to specific joints. In Fig. \ref{fig:viz_heatmap} we show the average per-class weights for the HMDB-1 test set. We see some interesting, distinctive patterns. There are some joints like \texttt{LHand} and \texttt{RHand} which are used by the model for most inputs. Since none of the actions requires an explicit focus on keypoints on the face, motion features for keypoints like eyes and ears tend to be suppressed by the model for most cases. We also see a couple of cases where the model focuses on a very few joints like sword exercise. This might be due to biases underlying the dataset wherein the person performing the action is often in a similar pose and relatively few joints are enough to recognize the action.
Further, they might not add new information than that given by other joints. We also see some usage of limbs that make intuitive sense. For example, as expected, the activity `eat' and `drink' predominantly depends on the upper limbs. 
The design of JMRN model allows us to see which joints were predominantly used by the model to infer activities. 

\noindent\textbf{Alternative Pose Extractors} Our approach is modular and allows use of any off-the-shelf pose extractor. This is unlike approaches like PA3D and SIP-Net which make use of features from pose extractor and require access to features from pose extractor. We also use a more accurate pose extractor AlphaPose~\cite{fang2017rmpe,li2018crowdpose,xiu2018poseflow} on the Charades dataset. JMRN with AlphaPose gives an improvement of +0.5 mAP over JMRN with OpenPose. This demonstrates our approach's modular nature and applicability to other pose extractors. For fair comparison with previous approaches we used OpenPose for the experiments in the paper. 

\noindent\textbf{Benefit of joint-reweighting} Apart from the benefit of visualizing the joints that the model focuses on, the joint-selection helps improve performance too. We see that using the concatenated per-joint feature without reweighting drops the performance by 1\% and 0.9 mAP on JHMDB and Charades respectively while having comparable performance on the HMDB dataset. 
\section{Conclusion}
We presented an approach to learn better representations for pose-based action recognition. The proposed JMRN extracts useful per-joint motion information before reweighing the information for collective inter-joint reasoning. Our novel joint-contrastive loss further improves the results. This leads to improved performance on the downstream task of action recognition. The proposed pose-aware data augmentation step, applies a cascade of global and group-wise jitter. Our proposed approaches lead to improvement over state-of-the-art on JHMDB, HMDB, Charades, AVA and Mimetics using pose heatmaps alone. Fusion with state-of-the-art RGB and flow-based model leads to further improvements showing complementary nature of the pose stream. 
Our proposed approach can be extended to explicitly handle missing joint information and people in the background. We leave these to future work.
\section{Acknowledgements}
The authors would like to thank Alex Hanson, Ketul Shah, Susmija Reddy and Anoop Cherian for their helpful suggestions. This research is partially supported by the ONR MURI Grant N00014-20-1-2787.
\newpage
\section{Supplementary}
In this supplementary material, we provide:
\setlist{nolistsep}
\begin{itemize}
    \item Additional implementation details for the approach.
    \item Additional details on modality fusion.
    \item Other variants for Joint-Contrastive loss.
    \item Analyses and experiments on data augmentations. 
    \item Experiments on clip selection.
\end{itemize}

\section{Additional Implementation Details}
We use Adam with a learning rate of 1E-4 for JHMDB and HMDB and 5E-4 for Charades. The learning rate is reduced by 0.1 times after the validation loss plateaus. Use of dropout after convolutional layers, as employed in \cite{choutas2018potion} was beneficial for the Charades dataset. None of our experiments use the background joint in an attempt to force the network to extract richer per-joint motion information. Following \cite{choutas2018potion}, we additionally use flip augmentation for all experiments. Experiments on HMDB and Charades use a batch size of 128 while experiments on the much smaller JHMDB used a batch size of 16. Further, for JHMDB we activate the contrastive loss after 50 epochs. 
The dataset specific hyper-parameters that we used are:

\setlist{nolistsep}
\textbf{J-HMDB:} 
\begin{itemize}
\itemsep0em 
\item $c_{dim}$ : 128
\item $\lambda$: 0.05
\item Epochs : 100
\item Training and evaluation on heatmaps of size $64\times 86$
\item $\beta=6,\gamma=3$
\item $\tau=0.3$
\end{itemize}

\setlist{nolistsep}
\textbf{HMDB:} 
\begin{itemize}
\itemsep0em 
    \item $c_{dim}$ : 64
    \item $\lambda$: 0.5
    \item Epochs : 200
    \item Training on random crop of size $64 \times 86$ and evaluation on center crop of size $64 \times 86$
    \item $\beta=4,\gamma=0$
    \item $\tau=0.1$ 
\end{itemize}

\setlist{nolistsep}
\textbf{Charades:} 
\begin{itemize}
\itemsep0em 
    \item $c_{dim}$ : 32
    \item $\lambda$: 0.5
    \item Epochs : 150
    \item Training on random crop of size $64 \times 64$ and evaluation on center crop of size $64 \times 64$
    \item $\beta=8,\gamma=4$
    \item $\tau=0.05$
\end{itemize}

\noindent\textbf{Experiments on the Mimetics dataset}
For experiments on Mimetics, we first train a model on Kinetics50 dataset which has the 50 classes common with Mimetics. Then, we evaluate this model on the Mimetics dataset. We used a batch size of 16, $c_{dim}=256$ and trained the model for 100 epochs,. We used Dropout with this model. SGD with momentum was used for training this model with lr of 0.01, momentum of 0.9. We found $\lambda=0$ to work well for this dataset. In addition, we found the use of Gumbel-Max trick~\cite{maddison2016concrete,jang2016categorical} in the weight-selector to be beneficial.  Patience of 2 was used for the scheduler. $\beta,\gamma = (3,0)$ was used to train the model. We generate heatmaps by parsing the dataset provided by \cite{yan2018spatial} and adding Gaussian blobs around the keypoints. Our model gave 44.4\% on Kinetics50 validation set. 

\noindent\textbf{Experiments on AVA dataset}
Since AVA is annotated with multi-person multi-label actions, the bounding box information is used to crop per-person pose information across the clip. These are then used to obtain the pose encoding. For a fair comparison, we used the bounding boxes provided by SlowFast \cite{feichtenhofer2019slowfast}. To account for multi-person context information, we also append the pose encoding of other people in the clip as a separate channel. We use a batch size of 64 and train the model with a learning rate of 5E-4. We used $\beta,\gamma=(3,0)$. Hyperparameters for joint-contrastive loss were set to $\tau,\lambda = 0.1,0.5$.  

\begin{figure}
  \includegraphics[width=0.3\textwidth,scale=0.3]{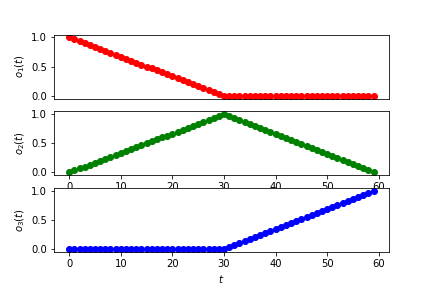}
  \centering
  \caption{Channel-time encoding $o[t]$ used to encode the pose heatmaps. We use the same $o[t]$ as \cite{choutas2018potion} with three channels. This example is for a video of sixty frames.}
  \vspace{-0.1in}
  \label{fig:aggregation_functions}
\end{figure}

\noindent\textbf{Additional baseline details}
Since an official implementation of \cite{choutas2018potion} is not publicly available, we reproduce the results using our own implementation. As mentioned in the main paper, we make use of the same aggregation function for a fair comparison. For completeness, we include the aggregation function used  in Fig.~\ref{fig:aggregation_functions}. 

\noindent\textbf{Normalization:} Normalization is essential before using a representation as an input to a neural network. The max-over-channel strategy chosen in \cite{choutas2018potion} normalizes the input to lie between 0 and 1. We posit that it is also important to encode the time spent by each joint at a location. The max-over-channel strategy assigns $1.0$ as the maximum value in a channel irrespective of the actual duration that a joint spent at a particular spatial location which renders different representations incomparable. To solve this, we normalize by dividing the aggregated representation $p_j$  by the maximum possible value that a joint can accumulate at that location. Note that, in \cite{choutas2018potion} the authors mention that they tried this approach, but this did not improve their performance. However, we obtain considerable improvement upon using this normalization perhaps due to better modeling.  

\noindent\textbf{Architecture Details} Table \ref{tab: architecture_motion_extractor} shows the architecture of our motion extractor module which is shared among all joints to give $r_j$. An additional $1 \times 1$ convolution is used to generate $c_j$. Table \ref{tab: architecture_joint_selector} shows the architecture of our joint selector module which generates $w_j$ using $r_j$. The representation $c_j$ is weighed using the weights $w_j$ and passed to the classification module (Table \ref{tab: architecture_classification_module}) to generate the final classification scores. ReLU activation is used after all layers except for the last layer of joint selector which uses sigmoid. The projection layers used with contrastive loss are implemented as two MLP layers which take spatially pooled per-joint features. The output from the MLP is normalized to lie on a unit hypersphere. 
\\

\begin{table}
\centering
\caption{Architecture details of the Motion Extractor. This module extracts motion information from each joint separately. In our proposed approach JMRN, parameters of this module are shared amongst all joints except the initial BatchNorm layer.}
\begin{tabular}{ccc}  
\toprule
Layer & Kernel/Stride & Output of Layer \\
\midrule
BatchNorm & - & $3 \times 64 \times 86$ \\
Convolution      & 3/2 & $128 \times 32 \times 43$ \\
Convolution      & 3/1 & $128 \times 32 \times 43$ \\
Convolution      & 3/2 & $256 \times 16 \times 22$\\
Convolution      & 3/1 & $256 \times 16 \times 22$\\
\bottomrule
\label{tab: architecture_motion_extractor}
\end{tabular}
\end{table}

\begin{table}
\centering
\caption{Architecture of the joint-selector module. This module uses the motion representation from all joints to reweight the joints most useful for the task.}
\begin{tabular}{ccc}  
\toprule
Layer & Kernel/Stride & Output Size \\
\midrule
Convolution & 1/1 & $256 \times 16 \times 22$ \\
Average Pool & - & $256 \times 1 \times 1$ \\
FC      & 3/1 & $J$ \\
\bottomrule
\label{tab: architecture_joint_selector}
\end{tabular}
\end{table}

\begin{table}
\centering
\caption{Architecture of the classification module. The reweighed joints are the input to the module. This module performs the final classification.}
\begin{tabular}{ccc}  
\toprule
Layer & Kernel/Stride & Output Size \\
\midrule
Convolution & 1/1 & $512 \times 16 \times 22$ \\
Convolution & 3/2 & $512 \times 8 \times 11$ \\
BatchNorm   & - & $512 \times 8 \times 11$\\
Convolution & 3/1 & $512 \times 8 \times 11$ \\
BatchNorm   & - & $512 \times 8 \times 11$\\
Global Average Pool   & - & $512$ \\
FC   & - & $C$\\
\bottomrule
\label{tab: architecture_classification_module}
\end{tabular}
\end{table}

\section{Modality Fusion}
\noindent\textbf{Implementation Details} As discussed in the main paper, we propose to use a very simple learnt fusion scheme to combine the different modalities. For single label classification tasks, we learn a single scalar weighing parameter for each of the $M$ modalities to be fused and for a multi-label task (like Charades) with $C$ classes, we learn $M \times C$ 
parameters. We first use the pretrained RGB+Flow and pose models to extract per-modality logits for each clip. We then learn the parameters using the train set. During inference, the modalities are fused using the learnt logits. Specifically for a single label classification task, 
\begin{align}
    l_{combined} = \alpha_{RGB} l_{RGB} + \alpha_{flow} l_{flow} + \alpha_{pose} \l_{pose}
\end{align}
Where $l_{RGB}, l_{flow}, l_{pose} \in \mathbb{R}^C$ are logits extracted from the pretrained network and $\alpha_{RGB},\alpha_{flow}, \alpha_{pose}$ are optimized to minimize classification loss w.r.t $l_{combined}$. For JHMDB and HMDB experiments, we $L2$ normalize the extracted logits. We train all models for 100 epochs. The Adam optimizer was used for all experiments and we chose learning rates of 1e-3, 1e-4, 5e-4 for JHMDB, HMDB, Charades respectively. Since we work with pre-extracted logits and there are a very few parameters to learn, these experiments can be performed very quickly and can be trained on a CPU. While there are a lot of approaches proposed for RGB and flow, we primarily choose models which have a high performance and are publicly available. We expect similar improvement over other models too. 

\begin{figure*}[h]
    \centering
    \includegraphics[width=\linewidth]{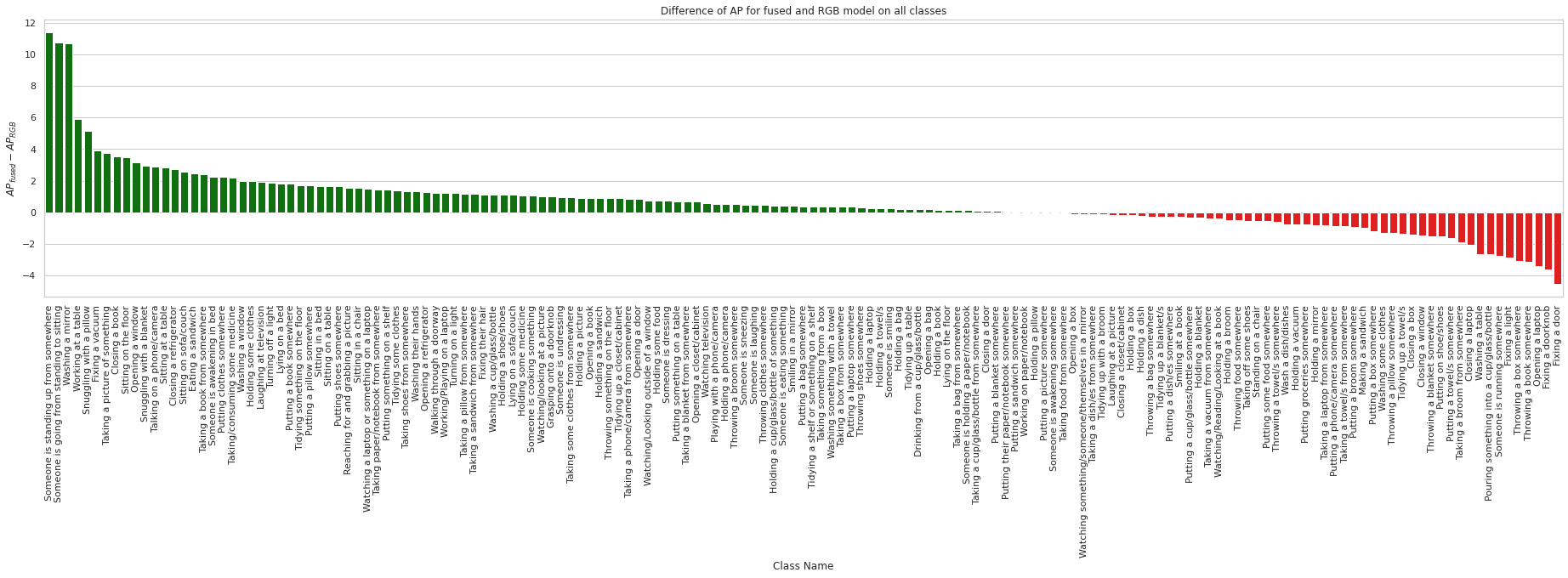}
    \caption{Per-class AP difference on Charades validation set when JMRN is combined with R101-NL-LFB\cite{wu2019long}. We see that most classes see an improvement in per-class AP scores justifying the claim that pose has complementary information. }
    \label{fig:fusion_charades}
    \centering
    \centering
    \includegraphics[width=0.7\linewidth]{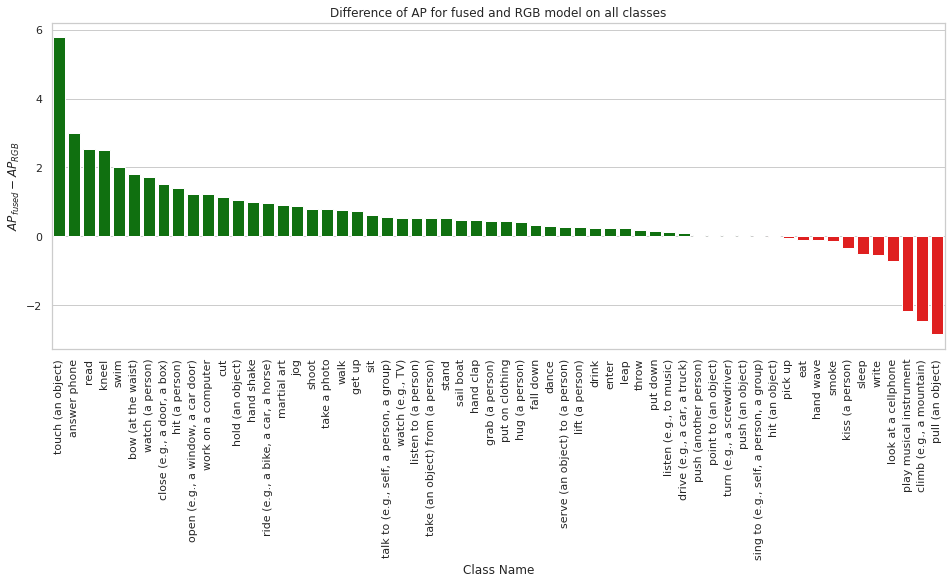}
    \caption{Per-class AP difference on AVA v2.1 validation set when JMRN is combined with SlowFast-R101-NL\cite{feichtenhofer2019slowfast}. }
    \label{fig:fusion_ava}
\end{figure*}

In Figures~\ref{fig:fusion_charades}, and \ref{fig:fusion_ava} we visualize the improvements that we obtain on per-class metrics when we perform fusion with recent state-of-the-art methods. 
We see considerable improvements on classes where human motion plays a key role and this improvement is also seen in the overall performance. We believe that training schemes which involve back-propagation through the pose and RGB/flow backbones like~\cite{joze2020mmtm} will lead to further gains and can avoid the drop that we see in a few classes during fusion. We leave this to future work. 

\section{Other variants for Joint-Contrastive loss}
Here we experiment with different variants of our joint-contrastive loss. We first experiment with a variant that does not use label information. Specifically, only the instance and its augmented version are considered as positives. With this approach every other instance in the batch is considered as a negative for loss calculation. Note that this approach is commonly used in the self-supervised learning setting. We refer to this variant as \texttt{only-aug}. As discussed in the paper, for multi-label problems we consider an instance as a positive if it shares any label with the anchor. An alternative approach is to weigh the positive examples by the number of labels that they have in common. We call this variant \texttt{multi-label-weighted}. We refer to the variant used in the main paper as \texttt{proposed}. The results are shown in Table~\ref{tab:contrastive_loss_ablation}. We see that the proposed variant uses the label information effectively during training and gives a better performance compared to the other variants. But, the other variants of joint-contrastive loss still outperform our model trained without any joint-contrastive loss thus showing the efficacy of the proposed approach. 

\begin{table*}
\centering
\small\renewcommand{\tabcolsep}{6pt}
\renewcommand{\arraystretch}{1.2}
\caption{Experiments with variants of joint-contrastive loss. We experiment with three variants of the joint-contrastive loss to train our model. We see that the \texttt{proposed} variant outperforms the alternatives. Further, we see that across all datasets the use of any of the variants improves performance over model trained without the joint-contrastive loss. This shows the effectiveness of the proposed loss in learning good representations.}
  \vspace{-0.05in}
\begin{tabular}{@{}llccc@{}}  
\toprule
\textbf{Model} & \textbf{Joint-Contrastive loss variant} & \textbf{JHMDB-1} & \textbf{HMDB-1} & \textbf{Charades}\\
\toprule
JMRN & None & 69.81  & 52.02 & 15.33  \\
\midrule
JMRN & \texttt{only-aug} & 69.01 & 53.13 & 15.91 \\
JMRN & \texttt{multi-label-weighted} & - & - & 15.97 \\
JMRN & \texttt{proposed}  & \textbf{71.08}  &  \textbf{54.05} & \textbf{16.2}\\
\bottomrule
\end{tabular}
\label{tab:contrastive_loss_ablation}
\vspace{-0.1in}
\end{table*}

\section{Analyses and experiments on data augmentation}

\begin{figure*}[t!]
    \centering
        \centering
            \includegraphics[width=0.206\linewidth]{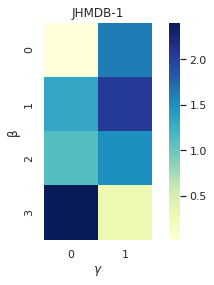}~~
            \includegraphics[width=0.36\linewidth]{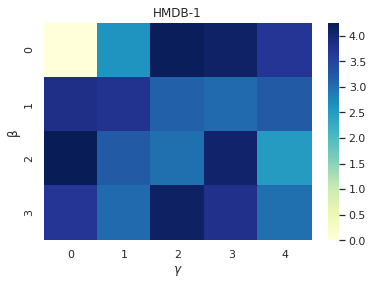}~~
            \includegraphics[width=0.36\linewidth]{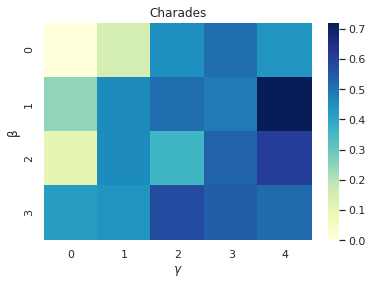}~~
    \caption{Effect of data augmentation hyperparameters $\beta,\gamma$. We visualize the improvement in accuracy compared to a model which does not use the proposed data augmentation step. We observe that even small values of $\beta,\gamma$ lead to a significant improvement which shows the effectiveness of the approach. Further, we notice that the performance is not very sensitive to the hyperparameter values and all values lead to an improvement.}
    \label{fig:beta_gamma_vis}
\end{figure*}

\textbf{Values of $\beta$ and $\gamma$.} 
\label{beta_gamma}
In Fig.~\ref{fig:beta_gamma_vis} we show the effect of $\beta$ and $\gamma$ on the performance of our model on the three datasets. It can be seen that almost all values of $\beta,\gamma$ lead to improvements over $(0,0)$ making it easier to use this method on other datasets without an extensive hyper-parameter search. Further, relatively small values of $\beta,\gamma$ are enough to give a considerable performance improvement. To evaluate the contribution of augmentation  scheme alone, we do not use contrastive loss in these experiments.

\noindent\textbf{Alternative Data Augmentation Technique.} We experimented with a few other data augmentation strategies. We first use a technique that was inspired by CutMix \cite{yun2019cutmix}. In the original paper, the authors generate new training examples by mixing patches from different images and changing the ground-truth labels based on the area of patches and the labels. We attempt to use this in our context. Specifically, during training, $P$ joints from a pose-representation are replaced with pose-representation from another video. The ground truth labels are appropriately changed and the network is trained with the modified ground truth. In our experiments, this strategy performed considerably worse than the proposed PAA. 

We also experimented with a pose-aware rotation augmentation. Specifically, we rotate the obtained representation about the center by a random amount. This can be seen to be a rotation equivalent to the global translation jitter in which the entire representation is moved by the same amount. Similar to the groupwise translation, we experiment with groupwise rotation too. We found the augmentation to be helpful and it leads to improvements over the baseline, though it performs slightly worse than translation based augmentation that was shown in the main paper. Combining the two augmentation in novel ways could lead to improved performance but we  leave that to future work.

It is to be noted that, unlike \cite{choutas2018potion}, which tried to randomly jitter each joint spatially, our approach of doing pose-aware augmentation retains the structure and hence leads to improvements. As observed in \cite{choutas2018potion}, we do not see any gain when we jitter or rotate each joint independently. 

\noindent\textbf{PAA with baseline} Our data augmentation step is general and can be applied to other models. To show this, we experimented with PAA applied to the baseline PoTion model. The results are shown in Table~\ref{tab:paa_with_baseline}. We see that the augmentation shows consistent improvement on the baseline but the results are still worse than our proposed model. Further, a joint-contrastive loss is not applicable to the baseline model since it does not extract per-joint motion features. 

\begin{table}
\centering
\small\renewcommand{\tabcolsep}{6pt}
\renewcommand{\arraystretch}{1.2}
\caption{PAA with baseline. The proposed augmentation step is not tied to JMRN but can give performance improvements to other methods. The baseline shows consistent improvement when used with PAA. Our proposed model still outperforms Baseline + PAA}
  \vspace{-0.05in}
  \resizebox{0.7\linewidth}{!}{ 
\begin{tabular}{@{}llccc@{}}  
\toprule
 & \textbf{Approach} & \textbf{JHMDB-1} & \textbf{HMDB-1} & \textbf{Charades}\\
\toprule
& PoTion & $59.44$  & $42.04$ & $13.54$ \\
 & + PAA &  $65.92$   & $49.76$  & $15.16$  \\ \midrule
& Ours  & \textbf{71.08}  &  \textbf{54.05} & \textbf{16.2}\\
\bottomrule
\end{tabular}}
\label{tab:paa_with_baseline}
\vspace{-0.1in}
\end{table}

\section{Filtering Clips for Action recognition using Pose Information}
\noindent\textbf{Pose guided clip selection}
In this section we explore application of pose trajectories for clip selection. Human trajectory information can contain significant information about the underlying activity. We posit that pose information can give crucial cues to select representative clips in a video. The intuition behind this is that the pose trajectory encodes the movement of all the joints along with redundancies and time spent at each location. Properties of this trajectory, like points of high or low curvature, time spent at a location, etc., can give cues to select clips that are discriminative for the task of action recognition. We next describe our approach for clip selection. 

We assume a trained clip classifier $f$ that takes as input clips $c_i \in \mathbb{R}^{3 \times H \times W \times T}$ for $i \in {1,\cdots ,N}$ from a video with $N$ clips and returns normalized logits $z_i$ that give the probability of occurrence of each class. The baseline model densely goes over each clip and uses a consensus function $g(z_1,\cdots,z_N)$ to give a single class-wise score for each video. Typically, \texttt{max} or \texttt{avg} is used as the consensus function. 

Inspired by \cite{korbar2019scsampler}, we generate `oracle' scores using the training data and the pre-trained clip classifier and then train a model to mimic the `oracle'. For clip selection experiments, we train using pose-representation extracted from clips of sixteen frames. Oracle for the trained clip classifier is first obtained using $\mathcal{O}_i =  \text{argtopK} ~(f_{y_i}(c_j))$ which selects the $K$ clips that give the highest confidence about the correct class. 
These Oracle clips are then used to generate ground truth labels: $\text{label}[i,j] = 1~\text{if } c_j \in \mathcal{O}_i$ and $0$ otherwise. Thus instead of selection being guided by RGB data, we use features extracted from JMRN. 

Next, we train a saliency ranker which ranks two clips during training time. In each batch, we sample an equal number of Oracle and non-oracle clips. During inference, we rank the clips according to the saliency given by the model for all clips in the video and select the topK.

\begin{figure*}[t!]
    \centering
        \centering
            \includegraphics[width=0.18\linewidth]{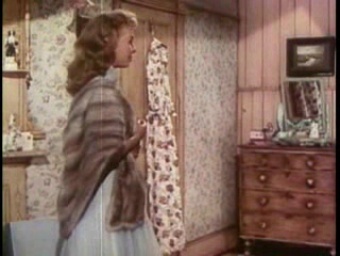}~~
            \includegraphics[width=0.18\linewidth]{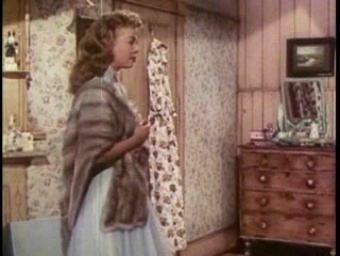}~~
            \includegraphics[width=0.18\linewidth]{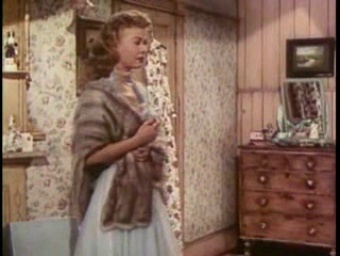}~~
            \includegraphics[width=0.18\linewidth]{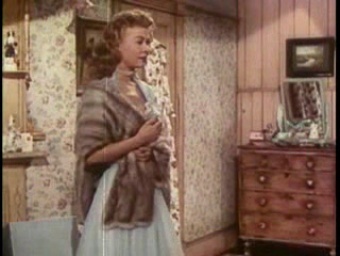}~~
            \includegraphics[width=0.18\linewidth]{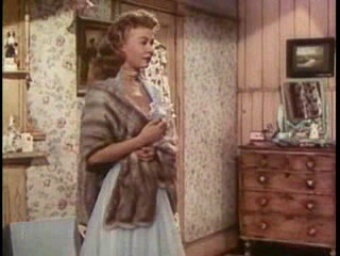}\\
            \includegraphics[width=0.18\linewidth]{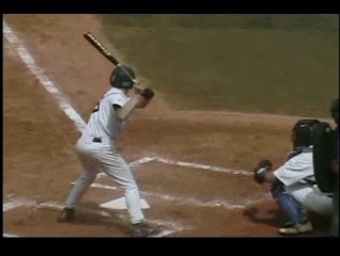}~~
            \includegraphics[width=0.18\linewidth]{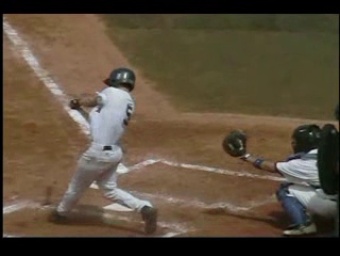}~~
            \includegraphics[width=0.18\linewidth]{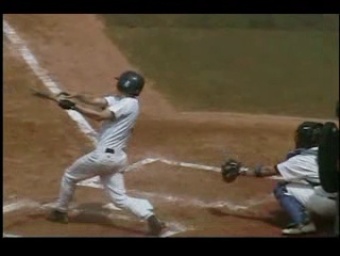}~~
            \includegraphics[width=0.18\linewidth]{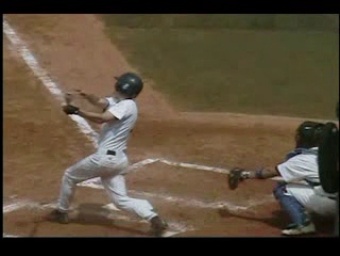}~~
            \includegraphics[width=0.18\linewidth]{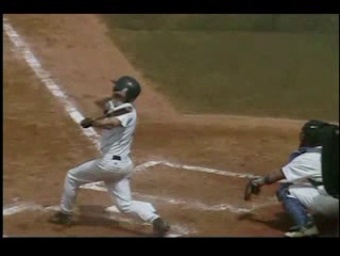}
    \caption{Here we show the middle frame corresponding to 5 most salient clips selected using our approach on the HMDB dataset. The upper row has frames from a video of `turn' whereas the second row is from a video of `swing baseball'. The frames are shown in order of their saliency scores. The selected clips are plausible for the underlying action.}
    \label{fig:clip_sel_vis}
\end{figure*}

In Table \ref{tab: clip_selection_all_splits} we show the results of clip selection using our approach on average of 3 splits on HMDB dataset. Using JMRN to extract pose features helps select salient clips and gives improvements over all baselines. We also show the middle frames corresponding to the 5 most salient clips in Fig. \ref{fig:clip_sel_vis}. The selected clips are plausible for the activity of `turn' and `swing baseball' respectively. 
\begin{table}
\centering
\small
\renewcommand{\arraystretch}{1.2}
\renewcommand{\tabcolsep}{6pt}
\caption{Clip selection comparison average of 3 splits on HMDB. Our model outperforms the dense clip selector baseline by 0.7\% while utilizing only 60\% of clips.}
\vspace{-0.05in}
\begin{tabular}{@{}lcc@{}}  
\toprule
Method & Clips used &  Accuracy(\%)\\
\toprule
Dense & 24 & 75.64 \\
Random & 14 & 74.72 \\
Uniform & 14 & 75.11 \\ 
Empirical & 14 & 74.96 \\
JMRN + PAA + TAN & 14 & \textbf{76.34} \\
\bottomrule
\label{tab: clip_selection_all_splits}
\end{tabular}
\vspace{-0.3in}
\end{table}

{\small
\bibliographystyle{ieee_fullname}
\bibliography{egbib}
}

\end{document}